\documentclass[runningheads]{llncs}
\usepackage[T1]{fontenc}
\usepackage{times}
\usepackage{graphicx}
\usepackage{url}
\usepackage{amsmath}
\usepackage[subpreambles=false]{standalone}
\usepackage{xspace}
\usepackage{amsfonts}
\usepackage{multirow}
\usepackage{color}
\usepackage{listings}
\definecolor{KeywordBlue}{cmyk}{0.88,0.77,0,0} %88,77,0,0
\definecolor{CommentGreen}{cmyk}{0.87,0.24,1.0,0.13} %87,24,100,13
\newcommand{\nolinkurl}[1]{texttt{#1}}

\begin{document}
\renewcommand{\thelstlisting}{\arabic{lstlisting}}

\title{Neural Machine Translating from\\ Natural Language to SPARQL}
%\author{}
\author{Xiaoyu Yin\inst{1} \and
Dagmar Gromann\inst{2} \and
Sebastian Rudolph\inst{1}}

%\institute{}
\institute{TU Dresden, N{\"o}thnitzerstrasse 46, 01187 Dresden, Germany \and
University of Vienna, Gymnasiumstrasse 50, 1190 Vienna, Austria
}
%\email{\{}}
%
\maketitle              % typeset the header of the contribution
\begin{abstract}
SPARQL is a highly powerful query language for an ever-growing number of Linked Data resources and Knowledge Graphs. Using it requires a certain familiarity with the entities in the domain to be queried as well as expertise in the language's syntax and semantics, none of which average human web users can be assumed to possess. To overcome this limitation, automatically translating natural language questions to SPARQL queries has been a vibrant field of research. However, to this date, the vast success of deep learning methods has not yet been fully propagated to this research problem. This paper contributes to filling this gap by evaluating the utilization of eight different Neural Machine Translation (NMT) models for the task of translating from natural language to the structured query language SPARQL. While highlighting the importance of high-quantity and high-quality datasets, the results show a dominance of a CNN-based architecture with a BLEU score of up to 98 and accuracy of up to 94\%.  
\end{abstract}
\section{Introduction}
\label{intro}
SPARQL~\cite{Harris2013} is the standard query language for retrieving and manipulating information contained in Resource Description Framework (RDF)~\cite{Herman2015rdf} graphs. In contrast to regular search engines, where natural language queries can be posed, using SPARQL requires knowledge about the entities in the domain to be queried as well as an understanding of the syntax and semantics of the language. For this reason, its use is generally limited to a group of Semantic Web experts proficient in the query language. To spread its accessibility to a wider audience, ways to automatically translate from natural language (NL) questions to SPARQL have been investigated over the last decades. This paper contributes to this line of work by evaluating the use of Neural Machine Translation (NMT) models to automatically translate NL questions to SPARQL queries. 

Approaches to produce SPARQL queries from a controlled NL \cite{ferre:hal-00943522} or using an intermediary format (e.g. \cite{Dubey2016}) have obtained good results on highly complex SPARQL queries. In view of the success of Neural Machine Translation (NMT) approaches, it comes as a surprise that very few such models have been utilized to address this problem~\cite{Cai2017,dong2016language,DBLP:journals/corr/abs-1709-00103}. Luz et al.~\cite{Luz2018} combine an attention-equipped LSTM encoder-decoder model with a probabilistic language model and obtain good results on the comparatively small Geo880 dataset. Soru et al.~\cite{Soru2018a,Soru2018} propose a dataset and two-layer LSTM model for this task, both of which are used as baselines in this paper. 

To advance the use of NMT models in this area, this paper presents a large-scale comparison of three distinct neural network architectures (Recurrent Neural Networks (RNNs), Convolutional Neural Networks (CNNs), and the Transformer model) with a total of eight models across three datasets (monument~\cite{Soru2018a},  Largescale Complex Question Answering Dataset (LC-QUAD)~\cite{trivedi2017lc}, and, as the largest of the three, DBpedia Neural Question Answering (DBNQA)~\cite{Soru2018dbnqa}). Results are evaluated with common NMT score BLEU, string-matching accuracy, and model perplexity. All datasets, model checkpoints, and evaluation graphs (perplexity and BLEU score) are available at \url{https://bit.ly/2CEu8uV}. 

This paper is structured as follows: We will first provide some preliminary details on SPARQL and NMT models. Section~\ref{method} details the encoding of SPARQL queries for usage in Sequence to Sequence models and the models that are used for this comparison. It also describes the evaluation metrics. Section~\ref{experiments} describes the datasets and experimental setup, before details on the results are provided. Prior to some concluding remarks, we discuss comparisons of datasets, models, and the limitations of this approach. 

\section{Preliminaries}
\label{preliminaries}
\subsection{SPARQL}
\label{sec:prelim_sparql}
SPAQRL~\cite{Harris2013} is a structured, syntactically and semantically defined language to query graphs in the RDF~\cite{Herman2015rdf} data format. A SPARQL query generally contains a set of triple patterns that, in contrast to RDF triples, might contain variables. A query consists of two parts: a \texttt{SELECT} operator that identifies the variables to appear in the query and a \texttt{WHERE} clause that provides the graph pattern to which a subgraph of the RDF data store is to be matched. Extensions with additional operators help to filter, limit, or group the results. Additionally, the language supports negation, aggregation, counting and several other features. Each resource is represented by an Internationalized Resource Identifier (IRI), generalizations of URIs, which may be abbreviated by prefixed names. An example of a SPARQL query is provided in Listing~\ref{lst:sparql}, where the first two lines exemplify a prefix declaration, which is required in order to resolve the IRIs in the query. All strings preceded by a question mark denote variables. The triple pattern in the \texttt{WHERE} clause of Listing~\ref{lst:sparql} returns a solution set, which in our example only contains \url{http://dbpedia.org/resource/Chess} as this is what both people in the query are known for and the only value matching the corresponding RDF graph.

\subsection{Neural Machine Translation}
\label{sec:nmt_models}
Neural Machine Translation (NMT) models are designed to automatically translate from one natural language to another. This section specifies some of the important developments in the field to justify our choice of models. 
One popular choice in NMT and Sequence to Sequence problems in general is the encoder-decoder architecture.
%Its two parts, the encoder and the decoder, are solely connected by way of sharing a compressed representation of the input sequence. The encoder receives an input sequence of tokens $n$ that it turns into a compressed representation $m$ where $m < n$ and $m$ is usually one high-dimensional vector called the context vector. The decoder receives the compressed representation and decodes it to an output sequence of tokens in $l$ where $m < l$. For instance, the encoder receives a sentence in German with five words, which are embedded in vector space and compressed to one high-dimensional vector. The decoder uses this vector to produce an output sequence of seven Frenc
The popularity of this setting is based on the fact that input and output sequence can be of variable length. It has also been shown to be superior to traditional phrase-based models in ease of extracting features, flexibility with regard to the configuration of models, and better accuracy~\cite{Wu2016}. 

Encoder and decoder can be implemented by means of any (and even different) neural network architectures, since they are solely connected by sharing a compressed representation of the input sequence. Sutskever et al.~\cite{Sutskever2014} implement encoder and decoder as a multilayer Long Short-Term Memory (LSTM) model and also experiment with reversing the order of the input sequence, where the reversed order improves the overall results. However, their architecture experiences a drop in performance with an increase in input sequence length. To address this issue, attention mechanisms have been proposed. Instead of condensing the entire input sequence into one context vector, attention takes the last hidden layer of the encoder and is trained to assign higher weights to those elements of the input sequence that are more important for a given time step than others, which are then combined to one context vector. This approach is usually referred to as global attention~\cite{Bahdanau2014} and comes with an increased training cost. For this reason, local attention has been proposed~\cite{Luong2015}, which only considers a subset of the whole input sequence in each time step. Since RNN-based architectures are time- and resource-intensive, alternative models based on Convolutional Neural Networks (CNNs) have been proposed~\cite{gehring2017convs2s}, which can parallelize of the process. One ground-breaking success in NMT was achieved with a transformer model~\cite{Vaswani2017} that, as further simplification of the architecture without recurrence or convolution, alternates fully connected layers and attention mechanisms. 
%Its base architecture has recently been extended by the idea of back-translation, which first translates the target sentences to a synthetic source sentence corpus and then translates forward in the actual bilingual parallel corpus from source to target \cite{edunov2018understanding}. 
%Such transformer models have also been tested within the context of traditional mixture models for machine translation with a shared encoder but different decoders \cite{shen2019mixture}. 
%In the past years, some novel paradigms for NMT have also emerged like the applications of Generative Adversarial Networks (GAN) in \cite{wu2017adversarial,yang2017improving}.

\subsection{Evaluation metrics}
\label{sec:eval_metrics}
\subsubsection{BLEU score.}
Human evaluations are usually time- and cost-intensive and to some degree subjective. %For this reason n-gram precision is very common as a measure to reflect on the adequacy and fluency of the translated sentence
%, usually combined with perplexity to evaluate a language model's behavior.  The number of consecutive sequences o $n$ words are counted in the reference translation and divided by the total number o n-grams in the candidate translation produced by the MT model. N-gram precision produces high scores if the candidate translation duplicates n-grams in the reference translation.  
As a more efficient alternative to evaluate machine translation outputs, Papineni et al.~\cite{Papineni2002} suggested the Bilingual Evaluation Understudy (BLEU) score. It counts the number of times an n-gram occurs in the reference translation(s), takes the maximum count of each n-gram, and then clips the count of the n-grams in the candidate translation to the maximum count in the reference. This helps avoiding distorted precision counts based on duplicates. However, there is a problem when the length $c$ of the candidate translation is too short, in which case the final precision is likely to be too high. To counteract this, a brevity penalty (BP) is introduced which is set to~1 if $c$ is larger than the maximal reference length $r$ and set to $\exp{(1-r/c)}$ otherwise. 
%\begin{align*}
%BP = \left\{
%\begin{array}{lc}
%1 & \text{if } c > r \\
%e^{(1-r/c)} & \text{if } c \leq r
%\end{array}\right.
%\end{align*}
%where $ c $ and $ r $ are the length of the candidate and reference translation. 
Next, a set $\{w_1,\ldots,w_N\}$ of positive weights with $ \sum_{n=1}^{N}w_n = 1$ is chosen to take the weighted geometric mean of the scores of the modified n-gram precision $ p_{n} $ with different n-gram sizes up to some maximum length $ N $. Consequently, the BLEU score is computed by
$$
\textit{BLEU} = \textit{BP}\cdot \exp\Big(\sum_{n=1}^{N}w_{n}\log p_{n}\Big).
$$
Experimentally, $ N{=}4 $ and uniform weights $ w_{n}{=}\frac{1}{N}$ have shown encouraging results~\cite{Papineni2002}.

\vspace{-1ex}
\subsubsection{Perplexity.}
Recall that for a target probability distribution $ p $ and an estimated probability distribution $ q $, the cross entropy $H(p,q)$ measuring their similarity is defined~by
\begin{align*}
H(p,q) = - \sum_{x} p(x) \log q(x),
\end{align*}
where $ x $ stands for the possible values in the distribution. The perplexity is then defined as the exponentiation of the cross entropy:
\begin{align*}
\textit{Perplexity}(p,q) = 2^{H(p,q)}.
\end{align*}
%It is possible to use $ e $ as the base instead of $ 2 $ and it depends on which one is used in the cross entropy.
%
Specifically for machine translation, the target distribution $ p $ is the one-hot encoding vector of the target vocabulary and $ q $ is obtained from the result of the output softmax layer. In practice, perplexity is calculated per batch or epoch where the cross entropy is averaged over all internal decoding steps beforehand. It has been shown in related research~\cite{luong2015deep,Wu2016} that perplexity is a good measure for MT and our results also confirm that source-conditioned perplexity strongly correlates with MT performance.

\section{Related Work}
\label{related}
The primary focus of our investigation has been on neural network models that can be used to map natural language statements to SPARQL expressions. Thus, we focus our related work on deep learning models but broaden the range to models that translate to other structured query languages. Approaches that do not use neural network architectures are  not considered here.

Dong et al. \cite{dong2016language} presented a method based on an encoder-decoder model with attention mechanism aimed at translating the input utterances to their logical forms with minimum domain knowledge. Moreover, they proposed another sequence-to-tree model that has a special decoder better able to capture the hierarchical structure of logical forms. Then, they tested their model on four different datasets and evaluated the results with accuracy as the metric.

Cai et al. \cite{Cai2017} proposed an enhanced encoder-decoder framework for the task of translating natural language to SQL, a query language similar to SPARQL but targeting relational databases instead of graph databases. They used not only BLEU \cite{Papineni2002}, but also query accuracy, tuple recall, and tuple precision for measuring the quality of output queries, and achieved good results.

Zhong et al. \cite{DBLP:journals/corr/abs-1709-00103} proposed a framework called \textit{Seq2SQL} that utilized an LSTM-based encoder-decoder architecture to translate NL questions to SQL. Input NL questions were augmented by adding the column names of the queried table. Correspondingly, the decoder was split into three components, predicting aggregation classifier, column names, and \texttt{WHERE} clause part of a SQL query, respectively. As opposed to conventional teacher forcing, the model was trained with reinforcement learning avoid that queries delivering correct results upon execution but not having exact string matches would be wrongly penalized. To address this issue in the evaluation, execution accuracy and logical form accuracy of the generated queries were measured.

Luz et al. \cite{Luz2018} also used an LSTM encoder-decoder model but the purpose is to encode natural language and decode into SPARQL. Furthermore, they employed a neural probabilistic language model to learn a word vector representation for SPARQL, and used the attention mechanism to associate a vocabulary mapping between natural language and SPARQL. For the experiment, they transformed the logical queries in the traditional Geo880 dataset into equivalent SPARQL form. In terms of evaluation, they adopted two metrics: accuracy and syntactic errors. While they obtained reasonable results to comparable approaches, they did not handle the out of vocabulary issue and lexical ambiguities. 

%\begin{figure}[h]
%\includegraphics[width=0.7\textwidth]{neural-sparql-machines}
%\centering
%\caption{The generator-learner-interpreter architecture of \textit{Neural SPARQL Machines} %\cite{Soru2018a}}
%\label{figure:nsm architecture}
%\end{figure}

Soru et al. \cite{Soru2018a,Soru2018} proposed a generator-learner-interpreter architecture, namely \textit{Neural SPARQL Machines} (NSpM) to translate any natural language expression to encoded forms of SPARQL queries. They designed templates with variables that can be filled with instances from certain kinds of concepts in the target knowledge base and generated pairs of natural language expression and SPARQL query accordingly. After encoding operators, brackets, and URIs contained in original SPARQL queries, the pairs were fed into a sequence to sequence learner model as the training data. The model was able to generalize to unseen NL sentences, and generate encoding sequences of SPARQL for the interpreter to decode.

\section{Methodology}
\label{method}
\subsection{SPARQL encoding}
Unlike natural language that can be easily tokenized, SPARQL queries are internally structured, combining elements of the query language with elements from the RDF data store and variables. Thus, our first step is to encode each query as a sequence. Following the encoding approach suggested by Soru et al.~\cite{Soru2018a}, URIs are abbreviated using their prefixes (if necessary) and concatenated with the entities using underscores; brackets, wildcards, and dots are replaced by their verbal description, and SPARQL operators are lower-cased and represented by a specified number of tokens. These operations can be implemented as a set of replacements and applying them turns an original SPARQL query to a final sequence which contains tokens that are only formed of characters and underscores. An example query is provided in Listing \ref{lst:sparql}, which is shown in Listing~\ref{lst:encoded}. After training, when an encoded form of a SPARQL query has been generated, it can be easily decoded back by reverse replacements.

\begin{lstlisting}[language=SPARQL,label=lst:sparql,caption=Example SPARQL query,basicstyle=\small\tt]
PREFIX dbr: <http://dbpedia.org/resource/>
PREFIX dbo: <http://dbpedia.org/ontology/>

SELECT DISTINCT ?uri 
WHERE { 
dbr:Sam_Loyd dbo:knownFor ?uri .
dbr:Eric_Schiller dbo:knownFor ?uri . }
\end{lstlisting}
\begin{lstlisting}[  breaklines=true,label=lst:encoded,caption=Encoding of SPARQL query from Listing~\ref{lst:sparql},basicstyle=\small\tt]
select distinct var_uri where brack_open dbr_Sam_Loyd 
dbo_knownFor  var_uri sep_dot dbr_Eric_Schiller dbo_knownFor 
var_uri sep_dot brack_close
\end{lstlisting}
%
%\begin{table}[h]
%\caption{Example for encoding of a SPARQL query to be used as input to an NMT model}
%\label{table:sparql_encoding}
%\begin{tabular}{c | p{10 cm}}
%\hline
%SPARQL:  & 
%\begin{lstlisting}[language=SPARQL]
%PREFIX dbr: <http://dbpedia.org/resource/>
%PREFIX dbo: <http://dbpedia.org/ontology/>
%
%SELECT DISTINCT ?uri 
%WHERE { 
%dbr:Sam_Loyd dbo:knownFor ?uri .
%dbr:Eric_Schiller dbo:knownFor ?uri . 
%}
%\end{lstlisting}  \\ \hline
%Encoded  &  
%{\small select distinct var\_uri where brack\_open dbr\_Sam\_Loyd dbo\_knownFor var\_uri sep\_dot dbr\_Eric\_Schiller dbo\_knownFor var\_uri sep\_dot brack\_close} \\ \hline
%\end{tabular} 
%\end{table}

\subsection{Description of tested NMT models}
\label{section:models}
We compare three types of network architectures with a total of eight individual NMT models. The three types are RNN-based, CNN-based, and self-attention models since those represented the best performing NMT architectures in the field at the time of the experiment without considering hybrid and ensemble methods. 
%At the time of this experiment, these models achieved the best results on MT translation tasks in their categories, however, hybrid and ensemble methods were not considered. 
Encoded SPARQL queries and natural language questions are fed to the network on a word-level. 

\vspace{-1ex}
\subsubsection{RNN-based models.}
Many variants of RNN-based models exist, since they were considered the most natural choice for Sequence to Sequence learning for a long time. The main differences can be found in the number of hidden layers, types of units, and other main architectural decisions. In a closely related approach, a two-layer LSTM network has been used to translate from natural language to SPARQL~\cite{Soru2018a}, denoted NSpM model. This model shall serve as our baseline.

A first test on the baseline model investigates the effect of attention. Global attention~\cite{Bahdanau2014} (NSpM+Att1) is compared to local attention~\cite{Luong2015} (NSpM+Att2) to evaluate their impact. Moreover, we adopt the model proposed by Luong et al.~\cite{Luong2015} (LSTM\_Luong) and the GNMT system proposed by Wu et al.~\cite{Wu2016}, both of which achieved state-of-the-art results on natural language translation benchmarks. The former is essentially a deeper LSTM with 4 layers and a local attention mechanism. For the GNMT we differentiate between a model with four layers (GNMT-4) and one with eight layers \mbox{(GNMT-8).} Starting from the third layer, the GNMT architecture uses residual connections to remedy the loss of information caused by the transgression through many layers. It essentially adds the input and the output of the LSTM cell  together and feeds the result as the input to the next layer. In addition, GNMT utilizes a bi-directional RNN on the first layer of the encoder that reads the input sequence from left to right and right to left and combines the two outputs before feeding them to the next layer. 

%To address the issue of rare words, GNMT system adopts the wordpiece model (WPM) that segments the words into wordpieces in the pre-processing stage and establishes a wordpiece vocabulary for the system. During decoding time, the decoder predicts wordpiece sequences which are subsequently converted back to word sequences. WPM is found to be beneficial for the translation accuracy and faster decoding speed \cite{Wu2016}. However, due to the reason that preserving the integration of DBpedia entities in the target vocabulary is needed, WPM is not used in our experiments. 

%RNN is the most natural choice for the machine translation task because of its sequential structure. Prior research generally confirms that certain RNN-based models are superior to traditional statistical machine translation methods in translation quality, in spite of some weaknesses on expensive computation and issues of rare words. 

\vspace{-1ex}
\subsubsection{CNN-based models.}
RNNs suffer from a decrease in performance for longer sequences, high computational costs,  and have issues with rare words. CNN-based models are capable of overcoming some of those issues. Long-range dependencies have shorter paths when the inputs are processed in a hierarchical multi-layer CNN as opposed to the chain structure of RNNs. A CNN is able to create representations for $ n $ continuous words in $ \mathcal{O}(\frac{n}{k}) $ convolutions with $ k $-width kernels, while an RNN needs $ \mathcal{O}(n) $. CNNs also allow for faster training since they permit parallelization over every element in a sequence, whereas the computations in RNNs are sequentially dependent on each other. On the other hand, the input needs to be padded before being fed into the model since CNNs can only process sequences of fixed length. An additional position encoding is required to provide the model with a sense of ordering in the elements being dealt with.

One of the first CNN-based approaches is the Convolutional Sequence to Sequence (ConvS2S)~\cite{gehring2017convs2s} model. It is still an encoder-decoder architecture with attention, where both consist of stacked convolutional blocks. Each block is composed of a single dimensional convolutional layer followed by a Gated Linear Unit (GLU). The input to the convolution block can either be the output of the previous layer or, for the bottom layer of the encoder, the combined word and position embeddings. Note that because each convolutional block can only receive a fixed number of inputs, the input of each block needs to be padded with zero vectors. There is a residual connection between the input to the convolutional layer and the output of the GLU. Compared to the encoder, each convolutional block in the decoder has one more attention module located after the output of GLU and before the residual connection. This type of attention is considered multi-step attention, since the higher layers each compute attention individually and have access to the information which elements the lower layers attended to. 
%We denote the output of $ L^{'} $ layers of the encoder convolutional block as $ \textbf{z} = (z^{1},...,z^{L^{'}}) $ where the output of $ l $-th layer is $ z^{l} = (z_{1}^{l},...,z_{m}^{l}) $ and $ l $-th layer is stacked above the $ l-1 $-th layer. We denote the output of $ i $-th convolution operation $ Y = [A\,B] \in \mathbb{R}^{2d} $ where $ A \in \mathbb{R}_{d} $ and $ B \in \mathbb{R}_{d} $ takes each half of $ Y $, then:
%\[ z_{i}^{l} = A \otimes \sigma(B) + z_{i}^{l-1} \]
%where $ \otimes $ is the point-wise multiplication and $ \sigma $ is the sigmoid function.

%\begin{figure}[h]
%\includegraphics[width=\textwidth]{images/convs2s_enc_dec.png}
%\centering
%\caption{The illustration of convolutional blocks in the encoder (left side) and decoder (right side) of ConvS2S model \cite{auliconvs2s}. }
%\label{figure:convs2s enc dec}
%\end{figure}

This architecture is able to parallelize the computations required during the training phase since the target elements are known beforehand and can be fed to the decoder once. However, during the inference stage where the target elements are not available, the computations in the decoder are still sequential. Nevertheless, full parallelization of the encoder is enough to make this model faster than most of its RNN rivals~\cite{gehring2017convs2s}.

\vspace{-1ex}
\subsubsection{Transformer models.}
Transformer models have turned out to be simple but effective without convolution or reccurence. Each layer of the encoder and decoder in this architecture has two sub-layers: a multi-head self-attention mechanism and a point-wise fully connected feed-forward network, with an additional, third multi-head attention layer in the decoder over the encoder stack~\cite{Vaswani2017}. In multi-head attention, the input is composed of three parts: queries, keys, and values which are all vectors. Each head performs a scaled dot-product attention that maps from queries and a set of key-value pairs to an output attention matrix, where the queries and all keys are computed first, then scaled, and finally put through a softmax function to obtain weights on the different positions of values. At the connection between encoder and decoder, the queries come from the previous decoder layer while the keys and values represent the outputs from the encoder. This model shows both quality and speed advantages and has achieved state-of-the-art results on multiple translation tasks.

\subsection{Evaluation}
To compare the outputs of the eight models in this paper, we employ a combination of evaluation metrics. First, we use the common NMT metric called BLEU score, which was detailed in Section~\ref{sec:eval_metrics}. While BLEU shows high correlations with human evaluations and low marginal computational costs~\cite{Papineni2002}, it does not account for word order. To this end, the exact string matching accuracy is computed as well as the F1 score based on individual syntactical elements of the candidate and target translation. In addition, we utilize perplexity to show the model's intrinsic behavior which represents the exponentiation of the cross entropy to the base of two. A better language model is one that assigns higher probabilities to the words that actually occur.

\section{Experimental setup}
\label{experiments}
\subsection{Datasets}
 \label{section:datasets}
To successfully train a neural machine translation model, a large-quantity bilingual parallel corpus is needed. Our setting also requires a large parallel corpus, however, with NL questions aligned with their corresponding SPARQL queries. In terms of NL, we only consider English in this evaluation. Some difficulties in creating such a dataset are that expertise in SPARQL is required, knowledge about the underlying database to be queried is necessary (e.g. ``located at'' is represented as \texttt{dbo:location} in DBpedia), and changes to the whole knowledge base (KB) can affect the validity of the dataset. The three datasets used were constructed by creating a list of template pairs (see Table~\ref{table:template pair}) with placeholders inside and then replacing the placeholders with extracted entities or predicates from the latest endpoint of an online KB. Due to this limitation and the complexity of SPARQL itself, only a subset of SPARQL operators are included in the target SPARQL queries of the involved datasets, which are:
 \texttt{SELECT}, \texttt{ASK}, \texttt{DISTINCT}, \texttt{WHERE}, \texttt{FILTER}, \texttt{ORDER BY}, \texttt{LIMIT}, \texttt{GROUP BY}, and \texttt{UNION}.

The Monument dataset is generated and used by the Neural SPARQL Machine~\cite{Soru2018a} system. It has 14,788 question-query pairs. The full vocabulary size is about 2,500 for English and 2,200 for SPARQL. The range of entities in this dataset is restricted to the instances of the specific class \texttt{dbo:Monument}, which is why we call it the Monument dataset. The data is generated from a list of manually crafted template pairs and related assistant SPARQL queries that can be executed directly on a DBpedia endpoint. For example, given a template pair in Table \ref{table:template pair},
\begin{table}[t]
\centering
\caption{A template pair in the Monument dataset}
\label{table:template pair}
\begin{tabular}{c|c}
Question template & Query template \\
\hline
Where is \texttt{<}A\texttt{>} ? & \begin{lstlisting}[language=SPARQL,basicstyle=\small\tt]
SELECT ?x
WHERE
{ <A> dbo:location ?x . }
\end{lstlisting}
\end{tabular}\\[2ex]
\end{table}
where \texttt{<}A\texttt{>} belongs to the class \texttt{dbo:Monument} in DBpedia, one can then retrieve a list of entities and their corresponding English labels to replace \texttt{<}A\texttt{>} by executing an assistant SPARQL query on a DBpedia endpoint. An example is shown in Table~\ref{table:example result generation}. It is reported~\cite{Soru2018a} that 38 manually annotated templates were used in generating the Monument dataset. For each query template, 600 examples were generated with the aforementioned method. 
%However, we found that out of these 38 template pairs there are some issues that have caused the whole generated dataset to be simpler than expected. Some template pairs have different question templates but the same query template or very similar-structured query templates, which means the translation model may favor generating some certain kinds of queries. Additionally, 
Some English templates are partial phrases instead of full sentence (e.g. latitude of \texttt{<}something\texttt{>}).

\begin{table}[t]
\centering
\caption{An example returned result and template instantiation from running an assistant query}
\label{table:example result generation}
\begin{tabular}{c|c}
?uri & \texttt{http://dbpedia.org/resource/Carew\_Cross}  \\
\hline
?label & ``Carew Cross"@en \\
\hline
Generated question & Where is Carew Cross ? \\
\hline
Generated query & \begin{lstlisting}[language=SPARQL,basicstyle=\small\tt]
SELECT ?x
WHERE
{ http://dbpedia.org/resource/Carew_Cross 
  dbo:location ?x . }
\end{lstlisting}
\end{tabular}
\end{table}

The Largescale Complex Question Answering Dataset (LC-QUAD)~\cite{trivedi2017lc} contains 5,000 pairs, in which about 7,000 English words and 5,000 SPARQL tokens are used. The SPARQL queries are for DBpedia. The goal of LC-QUAD is to provide a large dataset with complex questions where the complexity of a question depends on how many triples its intended SPARQL query contains. To produce this dataset, 38 unique templates as well as 5,042 entities and 615 predicates from DBpedia were involved in the generation workflow.
%Original
%The generation of data in LC-QUAD is different from that in the Monument dataset. Instead of allocating an executable SPARQL query for each English-SPARQL template pair to retrieve a list of entity instances, an entity seed list as well as a predicate whitelist are prepared beforehand. Next, each entity in the entity seed list is used as a seed to extract subgraphs from DBpedia through a generic SPARQL query. The triples in the subgraphs are then used to instantiate the SPARQL templates and the corresponding English templates,which are called Normalized Natural Question Templates (NNQT). After that, the instances of NNQT are examined and paraphrased through peer reviews to ensure grammatical correctness. An example in LC-QUAD is shown in Table \ref{table:lcquad generation}.
%Shortened version: 
In contrast to the Monument dataset generation method, LC-QUAD allocates executable SPARQL queries to each English-SPARQL template pair to retrieve a list of entity instances. Using a previously prepared entity seed lists and a predicate whitelist, DBpedia subgraphs are extracted using a generic SPARQL query. The triples in the subgraphs are utilized to instantiate the SPARQL and English templates. The final result is peer reviewed to ensure grammatical correctness. 
%
%\begin{table}[t]
%\centering
%\caption{An example question and its corresponding instantiation of the query template and instances in LC-QUAD generation~\cite{trivedi2017lc}.}
%\label{table:lcquad generation}
%\begin{tabular}{c p{10cm}}
%Template & SELECT ?uri WHERE \{ ?x e\_in\_to\_e\_in\_out e\_in\_out . ?x e\_in\_to\_e ?uri . \} \\
%\hline
%Query & SELECT ?uri WHERE \{ ?x dbp:league dbr:Turkish\_Handball\_Super\_League . ?x dbp:mascot ?uri . \} \\
%\hline
%Instances & What is the \texttt{<}mascot\texttt{>} of the \texttt{<}handball team\texttt{>} whose \texttt{<}league\texttt{>} is \texttt{<}Turkish Handball Super League\texttt{>}? \\
%\hline
%Question & What are the mascots of the teams participating in the turkish handball
%super league? \\
%\end{tabular}
%\end{table}

DBpedia Neural Question Answering (DBNQA)~\cite{Soru2018dbnqa} is the largest DBpedia-tar\-geting dataset we have found so far and a superset of the Monument dataset. It is also based on English and SPARQL pairs and contains 894,499 instances in total. In terms of vocabulary, it has about 131,000 words for English and 244,900 tokens for SPARQL without any reduction. DBNQA provides a remedy for some drawbacks of the previous two datasets. A large number of generic templates are extracted from the concrete examples of two existing datasets LC-QUAD and QALD-7-Train~\cite{usbeck20177th} by replacing the entities with placeholders. These templates can subsequently be used in the same approach as the one in the Monument dataset to generate a large dataset. A drawback that this dataset suffers from is memory shortages due to the large vocabulary, which nevertheless is considerably smaller than common natural language datasets making a comparison to those tasks more difficult. 

\subsection{Frameworks} 
\label{section:frameworks}
Two frameworks are used in this comparison due to their popularity in NMT, one based on TensorFlow and one on PyTorch. TensorFlow NMT\footnote{available at \url{https://github.com/tensorflow/nmt}}  provides a flexible implementation of the RNN-based NMT models. One can easily build and train a variety of RNN-based architectures by specifying the hyperparameters, e.g. number of encoder-decoder layers and type of attention, through designated Python program commands. This framework is used in our experiments for training, and testing five different models including three baseline 2-layer LSTMs, a 4-layer GNMT, and an 8-layer GNMT. The Facebook AI Research Sequence-to-Sequence Toolkit\footnote{available at \url{https://github.com/pytorch/fairseq}} (Fairseq)~\cite{gehring2017convs2s} is another framework that implements various Seq2Seq models based on PyTorch. Fairseq provides off-the-shelf models as well as packed hyperparameter to configure user experiments. We used it to train and test three models including the 4-layer LSTM with attention proposed by Luong et al.~\cite{Luong2015}, the ConvS2S, and the Transformer\footnote{All code will be made publicly available upon publication.}.

\subsection{Dataset splits and hyperparameters}
 \label{section:experimental setup}
We split each dataset in a ratio of 80\%-10\%-10\% for training, validation, and testing. Moreover, we do two splits on the Monument dataset. First, we split at a ratio \mbox{of 50\%-10\%-40\%} to evaluate the complexity of the dataset, and second, we use the splitting approach in~\cite{Soru2018a} to directly compare our results with NSpM. The latter split essentially fixes 100 examples for both validation and test set and keeps the rest for the training set. In summary, we have five experimental datasets, namely: Monument, Monument50, Monument80, LC-QUAD, and DBNQA. The hyperparameter settings for all models correspond to the original, best-performing natural language NMT settings and are as shown in Table~\ref{table:hyperparameters}, which details the number of layers for encoder and decoder, the number of hidden units, whether the model uses an attention mechanism, which optimizer it uses with which learning rate (lr), and the applied dropout rate. The final column \textit{Size} refers to maximum training steps inTensorFlow NMT (the first 5 lines) and to the maximum epochs in Fairseq (the last three lines). The training is based on cross entropy loss minimization. For the decoding, beam search of beam width 5 is used for all the experiments.  

\begin{table}[t]
\setlength{\tabcolsep}{0.3em}
\caption{Hyperparameter settings of this experiment}
\label{table:hyperparameters}
~\hfill\begin{tabular}{ | c | r | r @{\ \ \ }| c | c | l | c | r |}
\hline
\textbf{Name} & \textbf{Layers} & \textbf{~H. Units} & \textbf{Attention} & \textbf{Optimizer} & \textbf{lr} & \textbf{Dropout} & \textbf{Size} \\ \hline
NSpM\footnote{from \textit{Neural SPARQL Machines} (available at \url{https://github.com/AKSW/NSpM})} & 2 & 128 & - & SGD & 1 & 0.2 &50,000 \\
NSpM + Att1 & 2 & 128 & yes & SGD & 1 & 0.2 & 50,000 \\
NSpM + Att2 & 2 & 128 & yes & SGD & 1& 0.2 & 50,000 \\
GNMT-4 & 4 & 1,024 & yes & SGD & 1 & 0.2 & 30,000 \\
GNMT-8 & 8 & 1,024 & yes  & SGD & 1 & 0.2 & 30,000 \\
LSTM\_Luong & 4 & 1,000 & yes  & Adam & 0.001 & 0.3 & 500 \\
ConvS2S & 15 & 512*\hspace{-1.2ex} & yes  & SGD & 0.5 & 0.2 & 500 \\
Transformer & 6 & 1,024 & yes & Adam & 0.0005 & 0.3 & 500 \\  \hline
\end{tabular}\hfill~\\[1ex]
\begin{footnotesize}
* first 9 layers: 512 kernel width 3; next 4 layers: 1,024 kernel width 3, final two layers: 2,048 kernel width 1 
\end{footnotesize}
\end{table}
%Table \ref{table:hyperparameters} details the number of layers for encoder and decoder, the number of hidden units, whether the model uses an attention mechanism, which optimizer it uses with which learning rate (lr), and the applied dropout rate. The final column $Size$ refers to maximum training steps in Fairseq (the first 4 lines) and to the maximum epochs in TensorFlow NMT (the last three lines). The training is based on cross entropy loss minimization. For the decoding, beam search of beam width 5 is used for all the experiments. 
%Full details of the experiments are available online\footnote{\url{https://github.com/xiaoyuin/tntspa}}.

\subsection{Runtime environment} 
\label{section:runtime environment}
Given that we have 40 different experiments where each experiment consumes various amounts of memory depending on the size of its model and dataset, we assigned them to three GPUs with memory capacity from small to large running on a High Performance Computing (HPC) server, the configurations are listed in Table~\ref{table:hpc gpus}. In terms of assignment of the resources to the individual models, we used only the small configuration for all models and datasets with the exception of the GNMT-8 dataset that always utilized the medium setting except for DBNQA, where it used the large setting. On DBNQA the first three models in Table~\ref{table:hyperparameters} were run in the medium and all others in the large configuration. All of the training was completed using Linux with Python 3.6.4, TensorFlow 1.8.0, and PyTorch 0.4.1 installed. 

\begin{table}[t]
\caption{Three hardware configurations on the HPC server used in this paper}
\label{table:hpc gpus}
\centering
\small
\begin{tabular}{|c |@{\ } l @{\ }|@{\ } l @{\ }|@{\ } l @{\ }|}
\hline
& GPU Small & GPU Medium & GPU Large \\ \hline \\[-2.3ex]
CPU & Intel\textsuperscript{\textregistered} Xeon\textsuperscript{\textregistered} CPU & Intel\textsuperscript{\textregistered} Xeon\textsuperscript{\textregistered} CPU & POWER9  \\  
    & E5-2450 @ 2.10GHz & E5-2680 @ 2.50GHz & \\  \hline
RAM & 24 GB & 16 GB & 192 GB (approximately) \\ \hline
Cores & 8 & 6 & 32 \\ \hline
GPU & NVIDIA\textsuperscript{\textregistered} Tesla\textsuperscript{\textregistered}  
    & NVIDIA\textsuperscript{\textregistered} Tesla\textsuperscript{\textregistered} 
    & NVIDIA\textsuperscript{\textregistered} Tesla\textsuperscript{\textregistered} \\     & K20Xm & K80 & V100-SXM2 \\ \hline
~GPU RAM~ & 6 GB & 12 GB & 32 GB \\
\hline
\end{tabular}
\end{table}

\section{Results}
\label{results}
In order to reflect on how well our models have been trained on different datasets, we report the perplexity for each experiment along with the training steps (from TensorFlow NMT) or epochs (from Fairseq). During the training, we stored the model checkpoints of the step or epoch which led to the best\footnote{Depending on the support of the frameworks, we stored the one with best validation BLEU in TensorFlow NMT and the one with best validation loss in Fairseq.} performance on the validation set in a model checkpoint file which is later used to perform decoding on the test set and report the BLEU scores. This section describes the statistical results of the 40 experiments, while their analysis and implications are discussed in Section~\ref{discussion}.

\subsection{Perplexities} 
\label{subsection:perplexity}
All final perplexity scores per model and dataset are reported in Table~\ref{table:ppl}, where \textit{T} represents training scores and \textit{V} refers to validation scores. In \textit{Mon}, which represents the Monument dataset, a full epoch for Fairseq is approximately equivalent to 120 steps for TensorFlow NMT. All of the complexities come close to one for this dataset. An unusual phenomenon can be observed for the Transformer model, where the validation perplexity becomes lower than the training perplexity. For the Monument80 dataset, around 100 steps in the TensorFlow NMT are equivalent to an epoch in the Fairseq models.With this dataset with less training examples, the validation perplexities are mostly higher than with the Monument dataset. Slight overfitting can be observed in the validation loss curve of the two NSpM with attention and the two GNMT models. Although the training size in the Monument50 dataset is nearly half cut, it appears that the performance on the validation set is not much affected, especially for LSTM\_Luong, ConvS2S, and Transformer which are all implemented in Fairseq, which is also reflected by their BLEU scores reported below. 

\begin{table}[t]
\centering
\setlength{\tabcolsep}{0.5em}
\caption{Perplexity scores for all models and all training and validation sets}
\label{table:ppl}
\begin{tabular}{ | r | r r | r r | r r |  r r | r r |} 
\hline
 & \multicolumn{2}{|c|}{Mon} & \multicolumn{2}{|c|}{Mon80} & \multicolumn{2}{|c|}{Mon50} & \multicolumn{2}{|c|}{LC-QUAD} & \multicolumn{2}{|c|}{DBNQA} \\ \hline
Models & T~~ & V~~ & T~~ & V~~ & T~~ & V~~ & T~~ & V~~ & T~~ & V~~ \\ \hline
NSpM & 1.00 & 1.09 &  1.00 & 1.21 & 1.00 & 1.29 & 1.00 &  16.46 & 2.05 & 2.32 \\ 
NSpM+Att1 & 1.01 & 1.16 & 1.00 & 1.44 & 1.00 & 1.62 &1.00 &  56.23 & 1.07 & 1.42  \\
NSpM+Att2 & 1.01 & 1.14 & 1.00 & 1.44 & 1.00 & 1.62 & 1.00 &  43.20 & 1.05 & 1.37 \\
GNMT-4 & 1.03 & 1.11 & 1.00 & 1.31 & 1.00 & 1.41 & 1.00 &   33.76 &1.74 & 2.24  \\
GNMT-8 & 1.04 & 1.15 & 1.01 & 1.32 & 1.00 & 1.59 & 1.01 & 229.96 &  2.13 & 2.43 \\
LSTM\_Luong & 1.11 & 1.19 & 1.11 & 1.24 & 1.11 & 1.26 & 1.12 & 4.92 & 1.90 & 2.15 \\
ConvS2S & 1.11 & 1.14 & 1.11 & 1.19 & 1.11 & 1.20 & 1.14 & 3.25 &  1.12 & 1.25  \\
Transformer & 1.13 & 1.09 &  1.12 & 1.14 & 1.14 & 1.17 &  1.16 & 3.15 & 2.21 & 3.34  \\ \hline
\end{tabular}
\end{table}

In the LC-QUAD dataset, we observed serious overfitting of all eight models. Most of the models have difficulty in providing low perplexity on the validation set, among which GNMT-8 performed the worst and ConvS2S the best. LC-QUAD is much smaller but more complicated compared to other datasets. It only takes 34 steps to finish training an epoch inside the Tensorflow NMT framework. Across all of the models, no evident overfitting is spotted for the DBNQA dataset. In a batch size of 128, an epoch of DBNQA takes nearly 5,600 training steps. Due to the large size of DBNQA, some models like NSpM and GNMT-8 have shown rather slow or incomplete convergence since the maximum training steps for them (50k and 30k) are just equivalent to a small number (10 and 6) of epochs. Nevertheless, all of the models have reached a perplexity of at least 2 on the validation set, where NSpM+Att1, NSpM+Att2, and ConvS2S have achieved lower than 2, which is especially reflected in their accuracy below.

\subsection{BLEU Scores}
\label{section:bleu} 
BLEU scores are reported on the validation \textit{V} and test \textit{T} set for each dataset for the best performing version of each model in Table~\ref{table:bleu}. ConvS2S outperforms all other models on all datasets. On all Monument datasets, the Fairseq models outperform the Tensorflow NMT models by a large margin. On the other two datasets, attention-equipped NSpM models perform equivalent to or even better than the Tensorflow NMT models. For the DBNQA dataset, they even come second place after ConvS2S. The low scores on the LC-QUAD dataset are also consistent with the high perplexity scores on this dataset. 

\begin{table}[t]
\centering
\setlength{\tabcolsep}{0.3em}
\caption{BLEU scores for all models and validation and test sets}
\label{table:bleu}
\begin{tabular}{| c | c c | c c | c c | c c | c c |}
\hline
 & \multicolumn{2}{|c|}{Mon} & \multicolumn{2}{c|}{Mon80} & \multicolumn{2}{c|}{Mon50} & \multicolumn{2}{c|}{LC-QUAD} & \multicolumn{2}{c|}{DBNQA} \\ \hline
Models & V & T & V & T & V & T & V & T & V & T  \\ \hline
NSpM &  80.43 & 80.28 & 87.55 & 87.03 &  85.19 & 85.54 & 43.91 & 43.50 & 65.89 & 65.92 \\
NSpM+Att1 & 80.36 & 80.58 &  87.82 & 87.34 &  85.98 & 86.17 &  52.68 & 50.13 &  89.87 & 89.87  \\
NSpM+Att2 &  80.88 & 80.03 & 87.99 & 87.37 & 86.60 & 86.52 & 53.03 & 50.86 & 91.51 & 91.50  \\
GNMT-4  & 80.12 & 79.53 &  85.94 & 85.39 & 82.92 & 83.01 & 43.69 & 42.71 &  69.65 & 69.61 \\
GNMT-8 & 79.30 & 79.07 &   84.94 & 84.14 & 80.35 & 80.76 & 44.32 & 43.91 & 68.43 & 68.41 \\
LSTM\_Luong & 92.39 & 91.67 & 96.35 & 96.12 & 94.05 & 94.75 &  52.43 & 51.06 &  77.64 & 77.67 \\
ConvS2S & \textbf{98.35} & \textbf{97.12} &  \textbf{96.74} & \textbf{96.47} & \textbf{96.44} & \textbf{96.62} &  \textbf{61.89} & \textbf{59.54} & \textbf{96.05} & \textbf{96.07}  \\
Transformer  & 95.25 & 95.31 & 95.16 & 94.87 & 93.80 & 93.92 &  58.99 & 57.43 & 68.68 & 68.82 \\ \hline
\end{tabular}
\end{table}

\subsection{Accuracy}
To take word order of the produced query into account, we count all produced queries that exactly match the target query and divide that count by the number of all queries in each subset of the datasets. The accuracy results and F1 scores presented in Table~\ref{table:acc} show that also for these measures, ConvS2S consistently outperforms all other models. Only once the LSTM\_Luong model produces a validation set accuracy equivalent to that of the best performer. These results also show the drastic problems with the LC-QUAD dataset, where most models fail to produce a single fully equivalent query and highlight the problems of some models on the DBNQA dataset. In the latter dataset, a positive effect of attention becomes evident with respect to the baseline model NSpM. 

\begin{table}[t]
\centering
\setlength{\tabcolsep}{0.4em}
\caption{Accuracy (in \%) of syntactically correct generated SPARQL queries $|$ F1 score}
\label{table:acc}
\begin{tabular}{|c | c c | c c | c c | c c | c c|}
\hline
 & \multicolumn{2}{|c|}{Mon} & \multicolumn{2}{c|}{Mon80} & \multicolumn{2}{c|}{Mon50} & \multicolumn{2}{c|}{LC-QUAD} & \multicolumn{2}{c|}{DBNQA} \\ \hline
Models & V & T & V & T & V & T & V & T & V & T  \\ \hline
NSpM & 71 $|$ 95 & 75 $|$ 93 & 75  $|$ 95 & 76  $|$ 95 & 82 $|$ 97 & 79 $|$ 96  & 0 $|$ 61  & 0 $|$ 61 &0  $|$ 77 & 0 $|$ 77\\
NSpM+Att1 &  71 $|$ 95 & 75 $|$ 93 & 77 $|$ 96 & 78 $|$ 96  & 83 $|$ 97 & 82 $|$ 97 & 	1 $|$ 68 & 1 $|$ 66 & 63 $|$ 93 & 63 $|$ 93 \\
NSpM+Att2 & 73 $|$ 96 & 74 $|$ 92 & 79  $|$ 97 & 78  $|$ 96 & 84 $|$ 97 & 81 $|$ 97 & 	1 $|$ 68 & 1 $|$ 67 & 69 $|$ 94 & 69 $|$ 94 \\
GNMT-4 & 70  $|$ 95 & 71  $|$ 92 & 67 $|$ 95 & 	68 $|$ 95 & 77 $|$ 96  & 75 $|$ 96 & 0 $|$ 62 & 0 $|$ 61 & 1 $|$ 84  & 1 $|$ 84 \\
GNMT-8 & 68 $|$ 95 & 73 $|$ 91 & 58 $|$ 94 & 60 $|$ 94 & 74 $|$ 96 & 71 $|$ 95 & 0 $|$ 65 & 0 $|$  64 & 0 $|$ 84 & 0  $|$ 84 \\
LSTM\_Luong & 75 $|$ 94 & 76 $|$ 94 & 82 $|$ 95 & 84 $|$ 96 & \textbf{90}  $|$ \textbf{98} & 89 $|$ 97 & 0 $|$ 68 & 0 $|$ 67 & 34 $|$ 82 & 34 $|$ 82 \\
ConvS2S & \textbf{94}  $|$ \textbf{99} & \textbf{95} $|$	\textbf{96} & \textbf{91} $|$	\textbf{98} & \textbf{90} $|$ \textbf{98} & \textbf{89} $|$ \textbf{98} & \textbf{90} $|$ \textbf{98} & \textbf{8} $|$ \textbf{74} & \textbf{8} $|$ \textbf{73} & \textbf{85}  $|$ \textbf{98} & \textbf{85} $|$ \textbf{97} \\ 
Transformer & 88 $|$ 98 & 91 $|$ 95 & 83 $|$ 96 & 84 $|$ 96 & 86 $|$ 92 & 84 $|$ 92 & 7 $|$ 71 & 4 $|$ 70 & 3 $|$ 79 & 3 $|$ 80 \\ \hline
\end{tabular}
\end{table}

\section{Discussion}
\label{discussion}
A comparison across the utilized datasets shows some explicit trends. One of the reasons for utilizing the Monument dataset is to allow for a direct comparison of the corresponding NSpM model~\cite{Soru2018a} with other NMT models. After training the NSpM, the results of a BLEU test score of 80 could be reproduced. Adding attention would be expected to outperform the baseline, which could not be observed in the BLEU scores. In the accuracy measure, attention mechanisms show their potential, since NSpM+Att2 (local attention) consistently outperforms the baseline. The GNMT models could not outperform this baseline. Changing the dataset split led to an increase in BLEU scores. We believe this is due to the increase of the number of examples in the validation and test set. Additionally, the difference between Monument80 and Monument50 in BLEU score is only 1-2 on average, which means that the models are still able to achieve a good performance after training from dramatically smaller proportions of the whole dataset. All of the above suggests that the monument dataset has little variation in sentence structures, which is expected from its generation methods (see Section~\ref{section:datasets}).

A dataset with higher variance is provided by LC-QUAD. However, due to its relatively small size and high complexity it seems not adequate for training deep learning models. All of the models experienced difficulties in training and showed low performances on the validation set. Especially in terms of accuracy most models fail to produce one fully equivalent and correctly ordered query. It could be the case that the data contained in the training set and validation set are somewhat distinct in types (i.e. using different templates) and the number of training samples is not enough to generalize the model to unseen data. In addition, even though the dataset was peer reviewed, some grammatical errors still exist and some questions  contain unusual punctuation that leads to undetectable incorrect tokenizations during vocabulary building (e.g. ``the u.n.i.t.y group" will be split into seven tokens ``the, u, n, i, t, y, group").

Given the above, we found DBNQA better suited to the task. It has a sufficient number of query types (i.e. templates) as well as volume for each template. From Table~\ref{table:ppl} and~\ref{table:bleu}, we found that the training of the models was normal overall and the results have shown the differences between the models. One unexpectedly low performance could be observed by the Transformer model. This might be attributed to a worse performance of self-attention with a very large vocabulary, which could not be observed with the other two attention-based models. However, when looking at accuracy, all but the attention-based and the ConvS2S models experience serious problems in producing a sequentially correctly ordered query. While we still believe that the DBNQA dataset is the best choice for training NMT models to translate from NL to SPARQL, the dataset also has obvious limitations. The vocabulary size is too large, which may be attributed to its generation method (see Section~\ref{section:datasets}) filling too many entities into the templates. We found it necessary to deal with this situation since the direct reduction of vocabulary size would definitely bring too many unknown tokens in the source and the target sentences. One of the possible solution is to adopt the generation approach proposed in LC-QUAD (see Section~\ref{section:datasets}) which somewhat restricts the number of involved entities and predicates in the beginning.

In terms of model performance, there is a clear winner in our comparison. From the results, it can be drawn that the ConvS2S model fits best for the specific task of translating natural language to SPARQL. The ConvS2S model outperformed other models  in BLEU scores on all of the datasets and achieved the fastest convergence and highest accuracy. One thing we did not expect is that the performance of GNMT models has been consistently below average on all of the datasets. Looking into the perplexity graphs, we found that GNMT-8 has mostly been the slowest in converging and suffered relatively large fluctuations. Although reducing the number of layers in GNMT-4 improved in converging speed a lot, it still struggles to give good results in BLEU scores and accuracy. We speculate that this might be due to the rather complicated architecture of GNMT, which to some extent may magnify the degree of overfitting. 

\begin{figure}[t]
\centering
\includegraphics[width=0.6\textwidth]{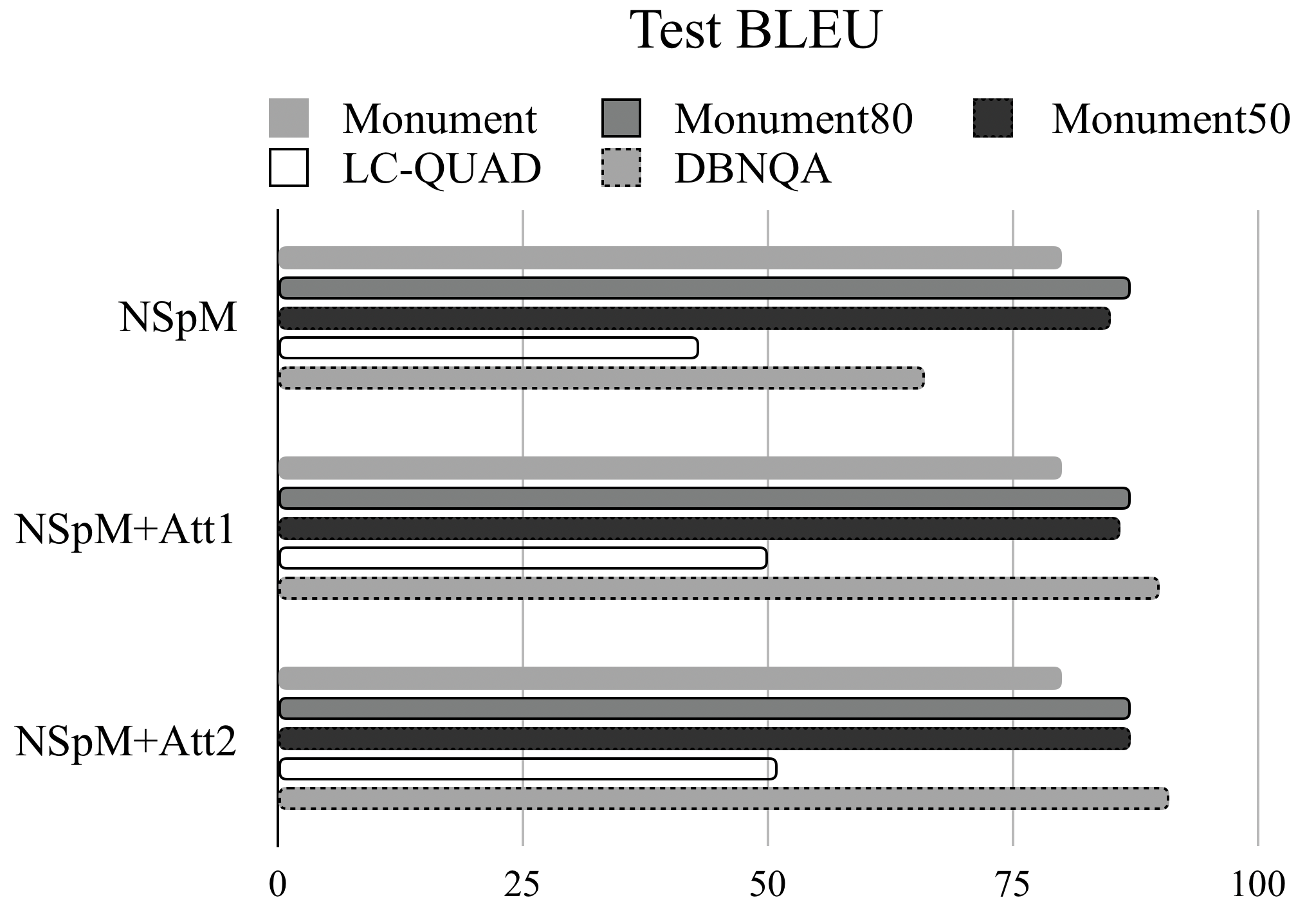}
\caption{The comparison between three NSpM models on test BLEU scores}
\label{fig:attention_comparison}
\end{figure}

Given the popularity of the attention mechanism, a direct comparison of its effect could be obtained by equipping the baseline model with two different types of attention. Figure~\ref{fig:attention_comparison} shows the comparison between three baseline RNN-based models on their test BLEU scores. The attention enhanced NSpM models generally performed equally to or better than the original NSpM. Furthermore, we found that NSpM+Att2 (local multiplicative attention) slightly performed better than NSpM+Att1 (global additive attention) on the LC-QUAD and DBNQA. Therefore, we believe that the attention mechanism can indeed boost the translation performance on the task of translating NL to SPARQL, similar to traditional NMT tasks.

Compared to their original task of translating between natural languages, the models performed higher on this task. The ConvS2S merely achieved a BLEU score of 26.43 at best on the WMT'14 English-German dataset~\cite{gehring2017convs2s}, while it is giving a highest of 97.12 and lowest of 59.54 in our experiments. This can be explained by the significant difference in the complexity of the datasets applied. The composition of SPARQL queries contained in our datasets does not vary much, e.g. they all start with the same headings such as \texttt{SELECT DISTINCT}.

In terms of evaluation metric, there is room for improvement. A clear correlation between perplexity and BLEU score could only be verified for the DBNQA dataset, but not for the other datasets. Experiments with accuracy revealed that the BLEU score is not entirely reliable when the word order of the output sequence is important. While models still obtained reasonable BLEU scores on the LC-QUAD and DBNQA datasets, some did not get a single query completely right. SPARQL syntax elements were generally produced correctly, however, problems with DBpedia entities could be observed, such as \texttt{dbr\_Radiant\_Silvergun} being replaced by \texttt{dbr\_Ella\_Fitzgerald}. 
%At other times they were close to the original, such as $dbp\_locationCity$ being changed to  $dbp\_location$. 
Most of the times, entities were replaced consistently with the same wrong element. While this combination of BLEU and accuracy provides a good estimation of model performance, it would be beneficial to devise an evaluation metric specific to this task or for translating to structured languages in general. To compare the performances of the models in more detail, it would also be necessary to fine-tune the hyperparameter settings to the task at hand specifically. This experiment, instead, utilized the default parameters provided for translating between natural languages.

\section{Conclusion and Future Work}
\label{conclusion}
In this comparative study on using NMT models for automatically translating from natural language to SPARQL queries, three representative neural network architectures were selected from which eight models were tested. All of these models were trained and tested on three different datasets and evaluated using perplexity, BLEU scores, and a simple accuracy measure to ensure the correct word order of the resulting SPARQL queries. As a result, we found that the ConvS2S model consistently, significantly outperformed all other models at a margin. This result could potentially be changed when fine-tuning the hyperparameters of the other models instead of using the default hyperparameters for translating between natural languages. Nevertheless, ConvS2S seems to be a solid choice for this task. In terms of dataset, the large and recent DBNQA dataset proved to be most adequate for the task at hand. 

Even though the DBNQA dataset allowed the models to improve their performance largely, it still suffers from limitations regarding the complexity of questions and queries and vocabulary size. A better suited dataset obtained by combining the generation method of LC-QUAD with the templates from the DBNQA may hold the promise of providing an even better NL to SPARQL dataset. Furthermore, an evaluation metric specifically targeted towards the evaluation of structured queries could strongly improve the comparison of deep learning models for this task. Additionally, an implementation of the models within the same framework and training environment would be desirable for a fully controlled comparison. Finally, these experiments were restricted to the English language. It would be interesting to see changes in the performance when utilizing a different natural language to generate SPARQL queries.  
\small
\bibliographystyle{splncs04}
\bibliography{references}

\end{document}